%% file: main.tex
\documentclass[lettersize,journal]{IEEEtran}
\usepackage{amsmath,amsfonts}
\usepackage{algorithmic}
\usepackage{array}
\usepackage[caption=false,font=normalsize,labelfont=sf,textfont=sf]{subfig}
\usepackage{textcomp}
\usepackage{stfloats}
\usepackage{url}
\usepackage{verbatim}
\usepackage{graphicx}
\def\BibTeX{{\rm B\kern-.05em{\sc i\kern-.025em b}\kern-.08em
    T\kern-.1667em\lower.7ex\hbox{E}\kern-.125emX}}
\usepackage{balance}

\usepackage{hyperref}
\usepackage{booktabs}

\begin{document}
\title{Joint Discovery of Object and Action Symbols through Effect Prediction for Robotic Manipulation Planning}

\author{
    Burcu Kilic\textsuperscript{1}\thanks{This work was supported by the Scientific and Technological Research Council of Turkey (TUBITAK) ARDEB 1001 Program (124E227) and by the INVERSE Project under Grant 101136067, funded by the European Union. \\ 
    \textsuperscript{1}Bogazici University, \textsuperscript{2}Ozyegin University, \textsuperscript{3}Osaka University.}, 
    Berke Kartal\textsuperscript{1}, 
    Fatih Dogangun\textsuperscript{1}, 
    Erhan Oztop\textsuperscript{2,3}, 
    Emre Ugur\textsuperscript{1}
}

\maketitle

\begin{abstract}
To perform complex manipulation planning, autonomous robots are required to abstract continuous, high-dimensional sensorimotor interactions into discrete object and action representations. Earlier work either categorized objects based on visual appearances, which fails to distinguish objects that appear similar but behave differently, or based on effects under interaction, but was limited to predefined actions. To address these limitations, we propose a model that jointly discovers high-level manipulation primitives and object categories through a binary bottleneck layer, trained to predict multi-modal outcomes, including object motion, contact, and force feedback, from random interaction data. Building on these discovered binary representations, we leverage a discrete planning method that uses intermediate steps in the predicted effect trajectory to enable partial action executions for precise low-level control. Additionally, we evaluate our framework's generalization capabilities on novel objects by assigning object categories through comparing a small number of interaction effects with the predicted effects of learned object symbols, enabling few-shot generalization based on behavior rather than visual similarity. We conduct experiments on tabletop repositioning and stacking tasks, and confirm that our effect-driven planning approach outperforms both a state-of-the-art method and a visual-based alternative in planning precision across seen and novel objects.
\end{abstract}


\input{introduction}
\input{related_work}
\input{method}
\input{experiments}

\input{conclusion}

\bibliographystyle{IEEEtran}
\bibliography{ref}

\end{document}

%% file: introduction.tex
\section{Introduction}

Autonomous robot development aims to build intelligent agents capable of learning through own experience, reasoning about their surroundings, and adapting to novel scenarios~\cite{asada2009cognitive}. However, a fundamental challenge is that reasoning over high-dimensional continuous sensorimotor spaces becomes unmanageable in long-horizon manipulation tasks. Operating in such complex sensorimotor spaces requires a layer of abstraction on both percepts and actions~\cite{konidaris2019necessity, konidaris2018skills}. For example, an infant that stacks multiple blocks on top of each other does not plan low-level joint movements directly from environmental pixels; rather, it thinks in terms of high-level intentions like picking a block and placing it on top of another one. 

Prior studies either manually constructed such abstractions~\cite{ugur2025} or used reconstruction-based clustering~\cite{asai2018classical}, ignoring the role of interaction. However, Sun~\cite{sun2000symbol} argued that the concepts of objects and skills emerge directly from the agent's interaction with the world and are linked to their goals and needs. Following this principle, an autonomous robot must discover its own representations by continuously exploring and manipulating objects, while learning to make sense of the physical dynamics. For this physical exploration to be effective, the robot's interactions cannot be limited to purely visual observations. Multi-modal feedback such as force and contact can convey complementary information to visual feedback about action outcomes, an observation also supported by developmental studies showing that young infants learn action properties through both visual and tactual interactions with objects~\cite{gibson1984development}. Motivated by this, our approach allows a robot to autonomously categorize objects and learn high-level manipulation skills directly from multi-modal physical interactions.

Through continuous physical interactions, children learn to categorize objects not only by static visual appearance, but primarily by the effects of actions performed on them~\cite{smith2005action}. This concept is defined by the term \textit{affordances}, first introduced in~\cite{gibson1977theory} as the action possibilities the environment provides to an agent relative to its own abilities. In this context, objects are learned not by visual appearance alone but by the actions the agent can perform on them and by the consequences of those actions~\cite{jamone2018affordances}. Following this, Ugur and Piater~\cite{ugur2015bottom} proposed a bottom-up approach to generate symbolic abstractions by grouping the similar effects of the robot's own continuous interaction with the objects. Ahmetoglu et al.~\cite{ahmetoglu2022deepsym} learned symbolic object affordances through a deep neural network architecture that predicts effects and showed that abstracting based on effects provides better planning performance than reconstruction-based clustering. Kilic et al.~\cite{kilic2025predictability} extended this method to learn object and parameterized action abstractions jointly. However, learning action primitives based solely on the final outcomes fails to capture the temporal structure of robotic trajectories. For example, if actions are categorized only by their end-effects, an agent cannot distinguish between pushing an object and picking-and-placing it to the exact same spot. This limitation motivates the use of temporally-extended actions and effects to capture the full temporal dynamics of an interaction, which also allows the use of intermediate points of a trajectory for precise planning. 

This deep understanding of physical dynamics is especially critical when encountering the unknown. Similar to an infant that applies familiar movements to new toys, an autonomous agent must be able to adapt its learned motor skills to handle novel settings with minimal interactions. Recent studies have utilized foundation models for one-shot object affordance grounding~\cite{tian2025oafford} or used geometric similarity to categorize new objects~\cite{ning2023where2explore}. However, these methods rely only on static appearance-based similarity and fail to capture the physical dynamics of interactions. Furthermore, foundation model-based approaches require computationally expensive training and lack data efficiency. We show that after learning functional object categories and high-level skills, generalizing to unseen objects based on behavioral similarity by matching multi-modal effects under minimal interactions allows more effective skill transfer than visual matching alone. 

In this study, we present a neuro-symbolic framework that enables learning high-level skills and object abstractions jointly from continuous random interaction data in a developmental setting and uses these learned concepts for planning and categorizing novel objects.
Our contributions are as follows:
\begin{itemize}
    \item A deep encoder-decoder network with a binary bottleneck that jointly discovers object and action symbols through a two-stage learning procedure optimized on multi-modal (spatial-haptic) effect trajectory predictions.
    \item A planning framework that builds a discrete effect library from the learned model and performs search with partial action executions, combined with symbol-conditioned low-level controllers and drift-based replanning for plan executions.
    \item A few-shot object generalization method, where novel objects are categorized according to behavioral similarity through effect matching, enabling manipulation planning on unseen objects.
    
\end{itemize}

We evaluate the proposed framework in a tabletop environment on manipulation tasks. Our results demonstrate that the proposed method, which uses the discovered symbolic actions, achieves significantly better planning performance compared to the Diffusion Policy~\cite{chi2024diffusionpolicy} baseline across all object types in both object repositioning and stacking tasks. We further conduct ablation studies to validate our design choices (a transformer-based action encoder and two-stage learning) and to analyze the impact of the action symbol dimension on planning performance. Moreover, we show that our effect-driven object categorization generalizes better to novel objects with a few demonstrations, outperforming the baselines that rely on visual conditioning of object depth maps.

%% file: related_work.tex
\section{Related Work}

It is necessary to discover object categories and high-level actions in order to perform discrete planning.~\cite{asai2018classical, asai2021learning, asai2022classical} used an autoencoder with a binary bottleneck trained on the reconstruction of states, allowing symbolic plan generation.~\cite{ahmetoglu2022deepsym, ahmetoglu2024discovering, ahmetoglu2025symbolic} abstracts object affordances by an effect predictor encoder-decoder deep neural network with a binary bottleneck layer; however, in these studies, predefined actions were used, and symbolic action discovery was not addressed.~\cite{kilic2025predictability} built upon a similar architecture and jointly discretized both actions and objects by spatial effect prediction in a curiosity-driven exploration setting. We perform a similar joint discretization that aims to obtain high-level skills and object categories, but we leverage partial action executions by predicting spatio-tactual effect trajectories and demonstrate that our framework can be beneficial in few-shot object generalization.

Learning the dynamics of the environment by effect prediction has been studied with different architectures.~\cite{tekden2024object} used Graph Neural Networks to model the effects of push actions on multi-part objects.~\cite{aktas2024multi} realized a model for single-action push and grasp effect prediction, and incorporated partial effects into their planning module.~\cite{kroemer2022search} integrated parameterized skills and learned continuous skill-effect models.~\cite{li2021planning} proposed a model-based planner over the learned latent action space.~\cite{girgin2024multiobject} learned multi-object affordances by effect prediction and utilized the network for multi-step planning; however, only a given place primitive was used as a skill. In addition to that, all of these architectures require model inference during planning, which is computationally heavy, whereas we create a discrete effect library to utilize in the planning.

Recently, generalization to novel objects by one- or few-shot demonstrations has been explored.~\cite{ning2023where2explore} categorized object affordances based on geometric similarity and generalized to novel object categories.~\cite{lorang2025fewshot} proposed a neuro-symbolic imitation learning framework; however did not explore autonomous discovery of skills and required a human labeler.~\cite{wang2022adaafford} realized a network to generate affordance maps for objects and performed active interactions to adapt a prior affordance map of a novel object.~\cite{tian2025oafford} used foundation models for object-to-object affordance grounding, which is useful in one-shot object generalization, and leverages LLMs for creating constraint functions during planning. Most of these methods are effective at identifying geometric similarity in novel objects, but we move beyond the visual affordances. We categorize objects based on their effects on the same actions; thus, we can differentiate objects that appear visually similar, but behave differently to the same applied action trajectory.

%% file: method.tex
\section{Method}
We propose a framework that autonomously discovers symbolic representations of objects and actions through predicting the effect of interactions, and utilizes these symbols for multi-step manipulation planning and few-shot generalization to novel objects. In this section, we first formulate the problem, then present the overview of the proposed framework, and detail its components in the subsequent sections.

\begin{figure*}
    \centering
    \includegraphics[width=\linewidth]{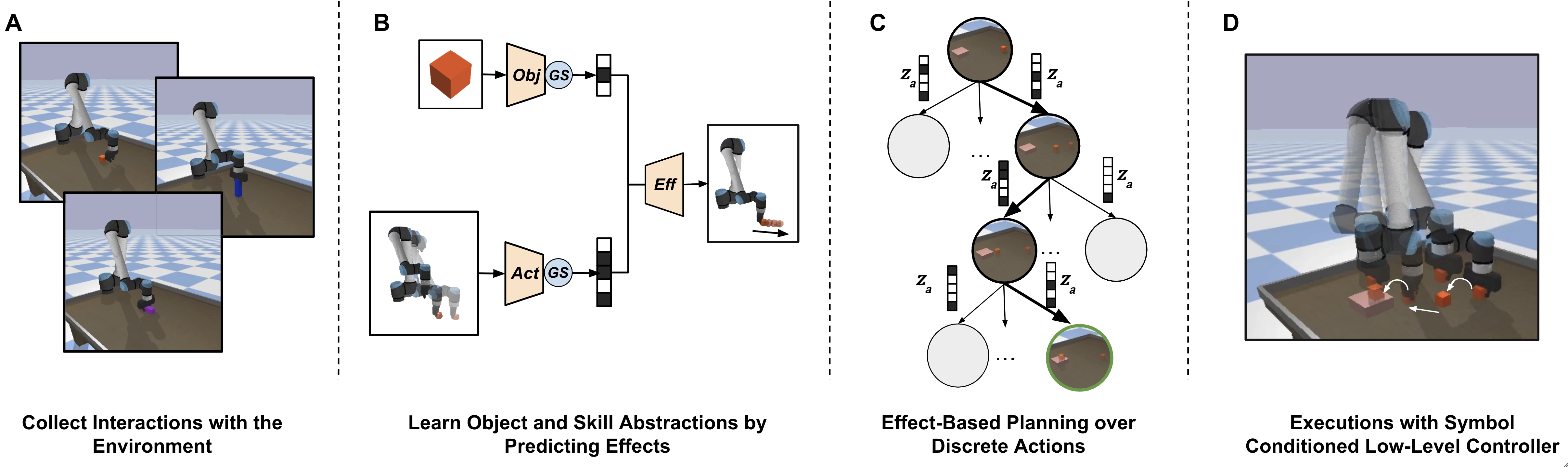}
    \caption{\textbf{Overview of the proposed framework.} \textbf{(A)} The agent explores the environment by interacting with objects, collecting depth images, joint angle trajectories, and multi-modal effect trajectories containing the position of the objects, force vectors, and contact feedback. \textbf{(B)} An encoder-decoder network, including object and action encoders with a Gumbel-Sigmoid (GS) activation function, which maps objects and actions to discrete symbols by learning to predict effect trajectories. \textbf{(C)} A discrete effect library is built from the discovered symbols, and an A* search over this effect library produces multi-step plans with partial action executions. \textbf{(D)} The planned symbolic actions are converted into executable joint-angle trajectories using a symbol-conditioned low-level controller.}
    \label{fig:method}
\end{figure*}

\subsection{Problem Definition}

We consider a robotic manipulation setting in which an agent interacts with objects on a tabletop environment and learns high-level object and action representations that enable multi-step discrete planning from its own exploration. In this section, we provide object, action, and effect formulations, and the objective for symbol discovery and planning. 

\subsubsection{Object Representation}

An object $o \in \mathbb{R}^{W\times H} $ is represented by its depth image with width $W=64$ and height $H=64$. 

\subsubsection{Action Representation}

An action $a \in \mathbb{R}^{T\times D}$ is a trajectory of 
robot's joints' angular positions over T timesteps, where D is the degree of freedom of the robot:
\begin{equation}
    a = \{\theta_1, \theta_2, \theta_3, ..., \theta_T\}, \quad \theta_t \in \mathbb{R}^D.
\end{equation}

\subsubsection{Effect Representation}

At each timestep $t \in \{1, 2, ..., T\}$ during the action execution, the following are recorded:

\begin{itemize}
    \item The absolute position of the object, $\mathbf{p}_t \in \mathbb{R}^3$.
    \item The force vector $\mathbf{f}_t \in \mathbb{R}^3$ measured from the force sensor on the robot's wrist.
    \item The contact feedback $c_t \in \mathbb{R}$ obtained from inside the gripper of the robot.
\end{itemize}

The effect trajectory is defined as the change in these quantities relative to the initial timestep:

\begin{equation}
    e = \{(\mathbf{p}_t-\mathbf{p}_0,\mathbf{f}_t-\mathbf{f}_0, c_t-c_0)\}_{t=1,\ldots,T}
\end{equation}

\subsubsection{Symbol Discovery} The aim is to learn mappings $\phi_o: \mathbb{R}^{W \times H} \rightarrow \{0,1\}^{s_o},$ and $ \phi_a: \mathbb{R}^{T \times D} \rightarrow \{0,1\}^{s_a}$ where $s_o$ and $s_a$ are the dimension of the object and action symbol, respectively; such that the learned discrete object and action symbols are instrumental about the effects, allowing discrete planning.

\subsubsection{Planning Problem} The aim is to find a symbolic action sequence $\pi$ that, when applied to a given object at position $p_0$, navigates it to the goal position $p_{goal}$. For the tabletop task, we measure planning performance by using Euclidean distance between the final position of the object and the goal $\|\mathbf{p}_{\text{final}} - \mathbf{p}_{\text{goal}}\|$. For the stacking task, we use a binary success criterion for whether the object is placed on top of a target object.

\subsection{Method Overview}

An overview of the proposed framework is illustrated in Fig.~\ref{fig:method}. The interaction data is collected through the robot's random exploration, including joint angle trajectories, depth images of objects, and multi-modal effect trajectories consisting of the position of the object, force, and contact feedback (Fig.~\ref{fig:method}-A). An encoder-decoder network is then trained to predict effect trajectories, yielding discrete action and object symbols in its latent space (Fig.~\ref{fig:method}-B). Using the discovered symbols, a discrete effect library is constructed over which A* search is performed to produce multi-step plans~(Fig.~\ref{fig:method}-C). Then, the symbolic actions in the plan are converted into joint-angle trajectories using symbol-conditioned low-level controllers~(Fig.~\ref{fig:method}-D). Once trained, the learned model and the effect library can additionally be leveraged for few-shot generalization to novel objects by matching their observed effects to the learned object symbols from only a few demonstrations.

\subsection{Discretization of Objects and Actions by Staged Learning of Effects}
\begin{figure*}
    \centering
    \includegraphics[width=\linewidth]{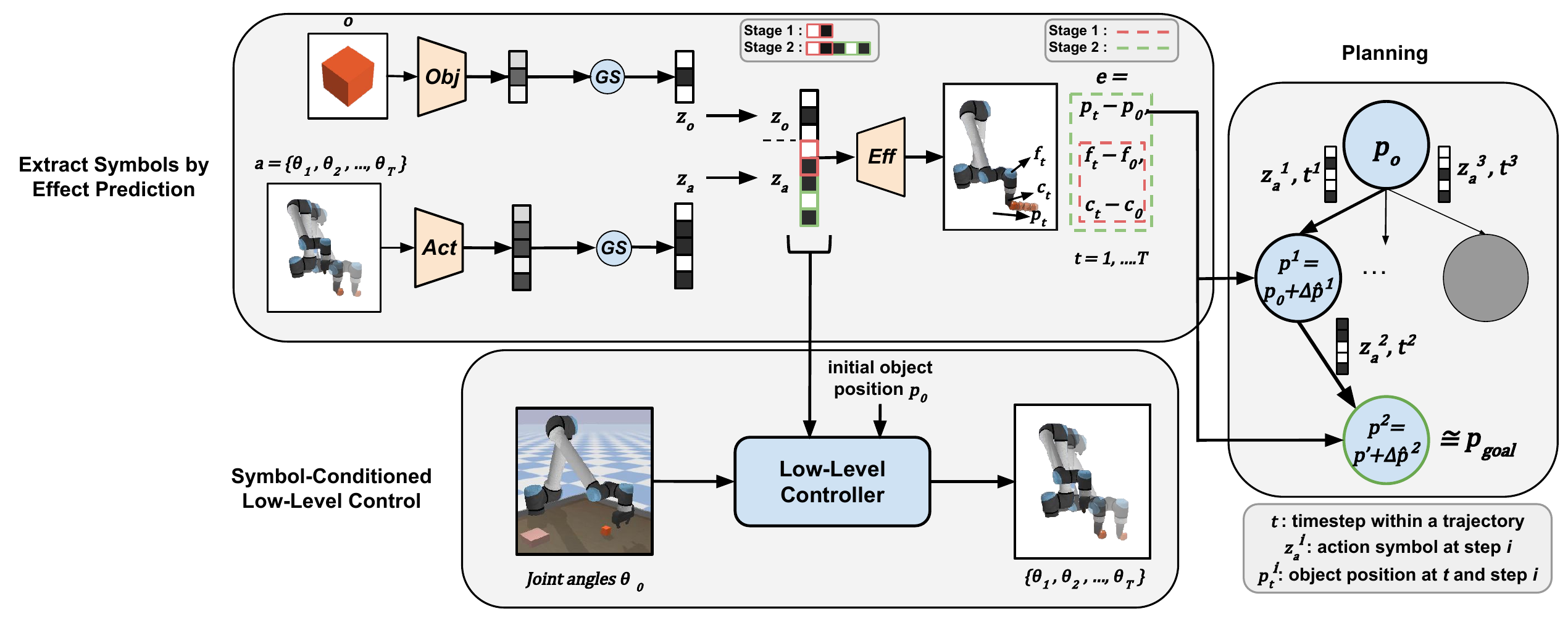}
    \caption{\textbf{Learning, planning, and execution pipeline of the proposed method.} Firstly, the effect prediction network takes joint angle trajectories and object depth maps from the interaction data, and produces discrete vectors by applying Gumbel-Sigmoid (GS) activation over the encoded inputs. In stage 1, the network predicts effect trajectories of force and contact feedback from a limited set of action symbol bits. In stage 2, the model uses all bits to predict all effects, including positional changes. After training is complete, object $z_o$ and action symbols $z_a$ are extracted. The low-level controller (CNMP) learns joint angle trajectories, conditioned on these object and action symbols. Lastly, a tree search planning is employed to reach a goal object position $p_{goal}$ given an initial position $p_0$. The planning is conducted on a discrete effect library that includes the predicted effects mappings from each tuple of object symbol, action symbol, and timestep ($z_o$, $z_a$, $t$).}
    \label{fig:detailed_method}
\end{figure*}

For discrete planning, we need discrete object and action representations that are predictive of effects. Reconstruction-based methods or unsupervised clustering can produce discrete categories. Still, they group by similarity in the input space, leaving the planner without a reliable link between symbols and outcomes. Therefore, object and action discretization must be learned jointly with effect prediction, since effects emerge only from the interaction of the two. We therefore propose a deep neural network that simultaneously predicts effect trajectories and produces symbolic abstractions through a binary bottleneck layer, which is shown in Figure~\ref{fig:detailed_method}.

The object encoder $\phi_o$, a convolutional neural network with a binary activation function, maps the given object depth image to a binary symbol $z_o$. The action encoder $\phi_a$, with a binary activation function, takes the joint angle trajectory over T timesteps and encodes it to a binary representation~$z_a$. Since action trajectories carry temporal structure in the form of sequential joint angles, we employ the Transformer architecture~\cite{vaswani2017attention} as the action encoder. Both encoders use Gumbel-Sigmoid (GS)~\cite{maddison2016concrete, jang2016categorical} as the binary activation function with temperature $Tp=1$:
\begin{equation}\label{eq:gumbel_sigmoid}
    GS(x) = \sigma(\frac{x+g}{Tp}), g=log(log(u_2)-log(u_1)+\epsilon)
\end{equation}

Here, the output is rounded to obtain binary symbols, and the reparameterization trick \cite{kingma2013auto} is used to pass gradients through both encoders during training. The binary symbols for object and action are concatenated as $r = (z_o || z_a)$, and the decoder $\psi$ predicts the effect trajectory $\hat{e}$ given this vector~$r$. 

We adopt a two-stage training procedure motivated by the finding that coarse action categories can be distinguished mainly from force and contact modalities, while 
directional action categories require spatial information. In the first stage, the model predicts only the force and contact feedback effects, using the full object symbol bits and fewer action symbol bits. This stage learns to differentiate high-level action primitives (e.g., distinguishing push from pick-and-place).
In the second stage, the model uses all symbol bits to predict the entire effect trajectory, including position, force, and contact feedback changes. This stage refines the action symbols to capture directional variations within each primitive (e.g., left push, right push, back pick-and-place).This procedure is also backed up by the developmental findings that infants first acquire coarse sensorimotor skills through tactile feedback, and later refine them into directional variants as visuomotor abilities improve~\cite{gibson1984development, oztop2004infant}. We find that this two-stage procedure yields empirically more stable training compared to learning all action categories simultaneously.
The model is trained with Mean Squared Error (MSE):
\begin{equation}\label{eq:mse_loss}
    \mathcal{L} = \frac{1}{T}\sum_{t=1}^T ||\hat{e}_t -e_t||^2
\end{equation}
After training, the model produces mappings for both objects $\phi_o(o)\xrightarrow{}z_o$ and action trajectories $\phi_a(a)\xrightarrow{}z_a$. To execute the learned symbolic actions on the robot, we need to invert the action mapping to generate continuous low-level trajectories from action symbols. 

\subsection{Symbol-Conditioned Low-Level Controller}

High-level action symbols enable efficient planning but cannot be directly executed on the robot. A low-level controller is required to generate executable joint angle trajectories conditioned on the discovered symbols.
We use Conditional Neural Movement Primitives (CNMP) \cite{Ugur-RSS-19} for this purpose, for two reasons. First, CNMP is a Learning from Demonstrations framework, so it can leverage the same interaction data used for training the effect prediction network without requiring additional trajectory collection.
Second, CNMP can be conditioned on task parameters, allowing the learned symbols to guide trajectory generation. 
A task or a demonstration is defined by the action to apply ($z_a$), the object it is applied to ($z_o$), and the object's initial position $(o_x, o_y, o_z)$. Therefore, the task parameter is constructed as:

\begin{equation}
    \gamma = [z_a,z_o,(o_x,o_y,o_z)]
\end{equation}

For each demonstration, the task parameter $\gamma$ is set using the symbols extracted from the trained effect-prediction model, along with the initial object location.

During training, CNMP predicts target joint angles $\theta_{target}$ from given observations $(t_{obs}, \theta_{obs}, \gamma)$, where both target $t_{target}$ and observation points $t_{obs}$ are uniformly sampled within a demonstration. In the forward pass, the CNMP encoder takes observations as input. A latent representation $L$ is obtained by applying average pooling over the encoder outputs. Then, a target timestep $t_{target}$ is concatenated with $L$ and queried to the decoder to predict the joint angles $\theta_{target}$ at that timestep. At inference time, a trajectory is generated by feeding initial joint angles with the task parameter to the encoder and querying the decoder with each target timestep. In this way, high-level action $z_a$ and object symbols $z_o$ are transformed into low-level actions that are executable by the manipulator robot. Please refer to \cite{Ugur-RSS-19} for more details.

\subsection{Planning with Partial Action Executions}
We utilize a search-based algorithm over discrete effects for planning. To this end, we first create an effect library that includes the positional changes exerted by the learned actions for each given object. We pass the object's depth image through the trained object encoder $\phi_o$ to find its symbol $z_o$. Then, for each discrete action symbol $z_a \in \{0,1\}^{s_a}$, we concatenate $z_a$ and $z_o$ and pass the resulting representation through the trained decoder $\psi$ to predict the effect trajectories $\hat{e}$. Since position changes are sufficient to perform spatial planning, we discard the contact and force feedback in the predicted effect, retaining only the position component of the effect $\hat{e}_{pos}$. Predicting the entire effect trajectory rather than only the final effect allows us to plan with partial action executions, where each intermediate timestep provides a distinct reachable state. 

Therefore, our constructed effect library consists of mappings from each combination of object symbol, action symbol, and timestep $z_o,z_a,t$ to a position effect $\hat{e}_{pos}^t$. Using this library, we can perform A* search over the space of discrete effects. Each node in the search tree corresponds to a 3D position of the object, with the root node corresponding to the initial position of the object. The edges represent transitions defined by the discrete effect predictions in the library. The A* search algorithm uses the Euclidean distance as the heuristic, and is bounded by maximum depth $d_{max}$. The search algorithm returns a plan when a state within an error margin $epsilon$ of $p_{goal}$ is reached. The resulting plan is a sequence of action symbols and timesteps: $\pi=\{(z_a^1,t^1),\ldots,(z_a^d,t^d)\}$ where $d$ is the length of a found plan, smaller than $d_{max}$.

To execute the generated discrete plans, we sequentially query the trained low-level controller for each step $k$, conditioned on $z_o$ and the $k$-th action symbol $z_a^k$. The resulting joint angle trajectory is truncated at the planned timestep $t^k$ to enable the partial execution.

To ensure the integrity of the execution, we implement a drift validation between steps. Before executing the step $k$, we compare the object's current position with the position predicted from the effect library. If the Euclidean distance between these positions exceeds a drift threshold $\delta$, we trigger a replanning with the same algorithm from the current position with remaining depth budget $d_{max}-k$. 

\subsection{Few-Shot Object Generalization}
\label{sec:few-shot}
To generalize novel objects, we categorize them based on behavioral similarity rather than visual appearance. We collect a small number of random demonstrations of a novel object, and use the trained action encoder $\phi_a$ to extract the action symbol $z_a$ for each demonstration. We then predict effects for each available object symbol $z_o$ combined with extracted action symbol, and compute the error between the predicted and observed effects. We finally assign the object symbol corresponding to the lowest mean prediction error across the demonstrations to the novel object, which enables planning on novel objects utilizing the existing effect library without retraining or large-scale data collection.

%% file: experiments.tex
\section{Experiments}
\begin{figure}[!t]
    \centering
    \includegraphics[width=\linewidth]{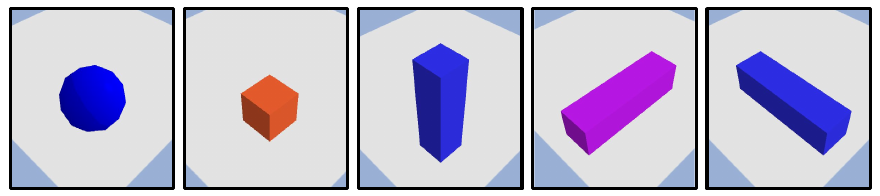}
    \caption{Objects used during the data collection. From left to right: Ball, Cube, Block Z, Block X, and Block Y.}
    \label{fig:seen_objects}
\end{figure}

In this section, we first explain our experimental setup, including the data collection procedure and training details. We then show the experiments we conducted to analyze the following:
\begin{itemize}
    \item Are the learned actions sufficiently diverse and precise for use in multi-step planning?~(Sec.~\ref{subsec:learned_effects})
    \item Does our transformer-based network with two-stage learning achieve high effect prediction accuracy?~(Sec.~\ref{subsec:exp_effect_pred})
    \item Does our framework outperform diffusion-based baseline across two different object manipulation tasks?~(Sec.~\ref{subsec:planning_performance})
    \item Is an effect-based approach stronger in few-shot object generalization than spatial approaches?~(Sec.~\ref{subsec:fewshot_exp})
\end{itemize}

\subsection{Experimental Setup}
\subsubsection{Data Collection}\label{subsec:data_collection} 

\begin{figure}[!b]
    \centering
    \includegraphics[width=\linewidth]{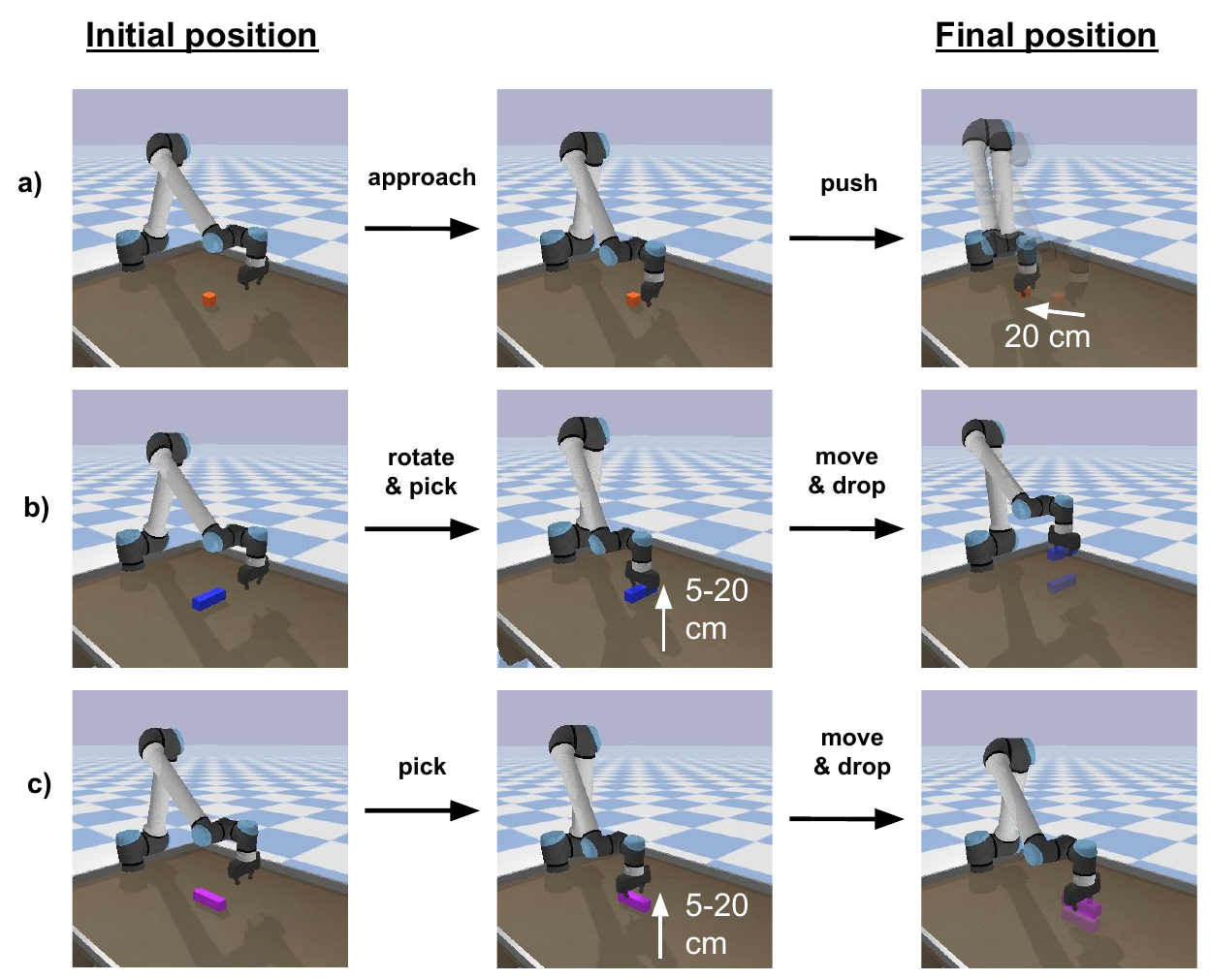}
    \caption{Data collection via random exploration. \textbf{(a) Push Action}: approach from a random angle and push the object. \textbf{(b) Pick \& Place (Y-aligned)}: approach while rotating the gripper, then grasp and lift the object. At last, move the object and drop it.  
    \textbf{(c) Pick \& Place (X-aligned)}: same as (b), but no gripper rotation is applied. Lift heights are uniformly sampled from 5-20 cm. Each move or push action moves the object 20 cm in a randomized direction.
    }
    \label{fig:data_collection}
\end{figure}

The experiments are conducted in PyBullet~\cite{coumans2021} tabletop simulation environment with a 6-DOF UR10e manipulator robot. Five different types of objects containing ball, cube, and long blocks with three different orientations, as shown in Fig.~\ref{fig:seen_objects}, are used for data collection. We name these long blocks Block X, Block Y, and Block Z by their alignment axis (e.g., Block Z is vertical). The data are collected through a random exploration, illustrated at Fig.~\ref{fig:data_collection}. An interaction starts with the robot approaching the object from a uniformly sampled direction. The robot then randomly executes either a push or a pick-and-place primitive on the object, with x-aligned or y-aligned grasp orientation. In the pushing primitive, the object is displaced by $20cm$ along a uniformly sampled direction on the table. In the pick-and-place primitive, the robot holds the object, lifts it to a uniformly sampled height $h \sim U[5, 20] cm$, and places it at a randomly sampled target location 20 cm away from the initial position. The two grasp orientations during the pick-and-place primitive enable the grasping of objects with different geometries. For example, a block wide along the X axis needs the X-aligned grasp to be lifted, but a cube works with both. The different grasps are also reflected in the force sensor readings recorded during interactions, allowing the effect-prediction model to differentiate objects based on their grasp affordances. We collected 1000 exploration data samples per object (5000 in total).

\subsubsection{Training}
The effect trajectory prediction model is trained on 5000 interaction samples consisting of joint angle trajectories, object depth maps, and effect trajectories. The model is trained for 600 epochs with a learning rate of $1\times10^{-4}$ using Adam optimizer \cite{kingma2014adam}. The object symbol dimension is set to $s_o = 3$, and the action symbol dimension is set to $ s_a = 5$ in the first training stage, allowing $2^5=32$ distinct actions to be learned.

After training the effect trajectory prediction model, we extract binary object and action symbols from the bottleneck layer using hard-rounded Gumbel-Sigmoid for both object and action encoders across all samples in the dataset. We then form the training data for the low-level controller using learned symbols along with the object initial positions and joint angle trajectories. We train the low-level controller CNMP, conditioned on object and action symbols and the object's initial position, to predict joint angle trajectories.

\subsubsection{Evaluated Planning Methods}
We compare the following methods in the planning experiments:
\begin{itemize}
    \item \textbf{Effect-Driven Planning Framework (Ours)}: 
    We use the effect-trajectory prediction network to extract action and object symbols and construct a discrete effect library. We employ A* search over the effect library, with partial action executions, to find plans, and then execute the symbolic actions of the plan using the trained CNMP as a low-level controller. 
    \item \textbf{Diffusion Policy}: 
     We train a Diffusion Policy conditioned on the goal position and the object's depth map. Since multi-step planning in our framework uses at most 4 action steps, we execute the Diffusion Policy at most 4 times sequentially for each test, and we condition each execution on the current position of the object. We consider an adaptive execution count for each test that is proportional to the distance of the given goal position.
    
\end{itemize}

\subsection{Generated Symbols}\label{subsec:learned_effects}

\begin{figure}[tb]
    \centering
    \includegraphics[width=\linewidth]{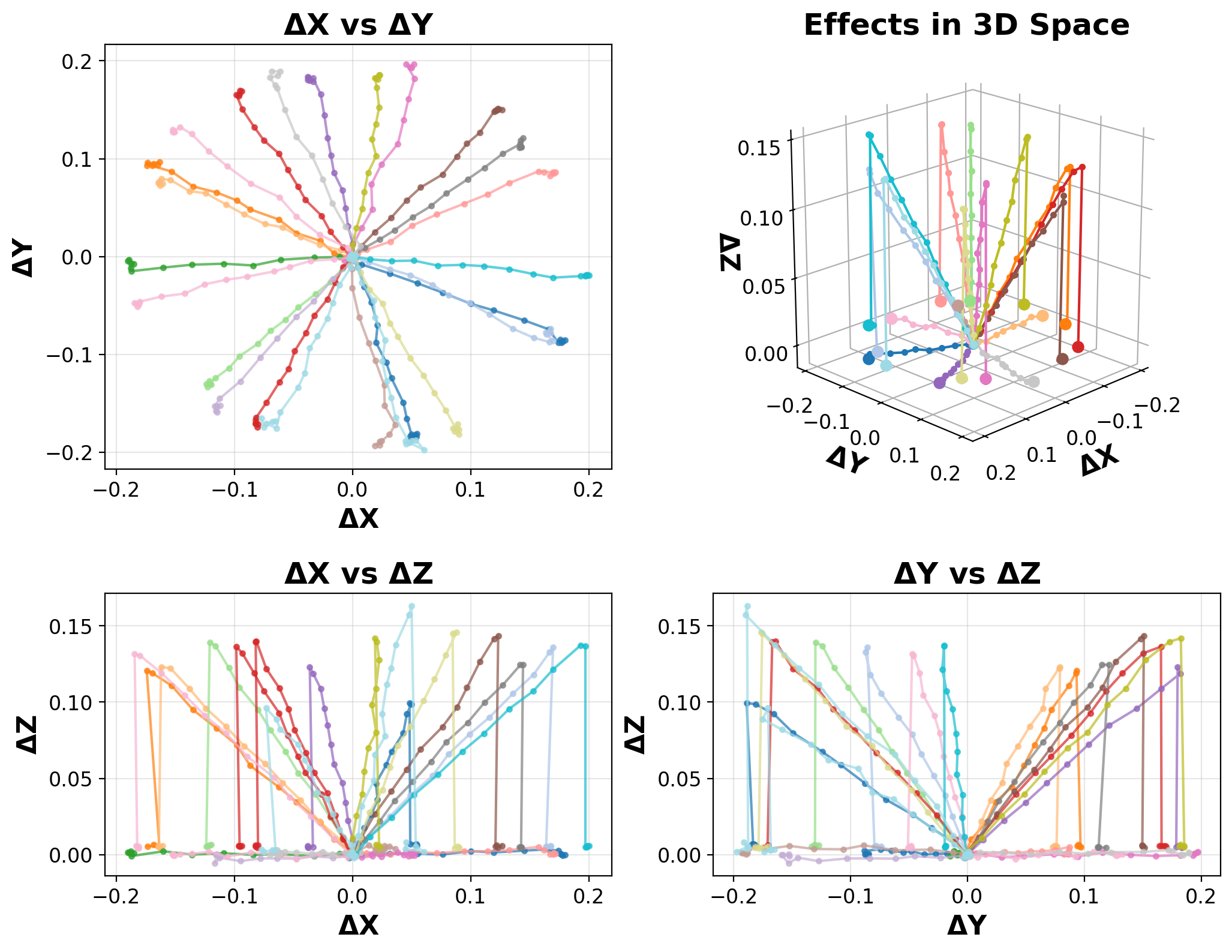}
    \caption{Visualization of the learned effect trajectories for the cube object. Each colored trajectory represents the predicted positional effect of a distinct learned action symbol. The 2D projections include all discovered actions, while some actions corresponding to less frequently encountered symbols are discarded from the 3D plot (on the top-right) for clarity.} 
    \label{fig:learned_effects}
\end{figure}

Fig.~\ref{fig:learned_effects} shows an example of the learned effect trajectories for the cube object, illustrating that the discovered action symbols capture a diverse set of manipulation primitives. The top-left projection illustrates the directional variety among the discovered primitives, while the bottom-row plots show the separation between push and pick-and-place primitives, where the pick-and-place primitives have a nonzero Z component indicating they involve lifting. The 3D plot on the top-right shows that the primitives corresponding to discovered action symbols cover the effect space with distinct directions and heights. This diverse and distributed set of action primitives is essential for planning, as they enable the planner to achieve precision in reaching arbitrary goal positions through multi-step planning.

\subsection{Effect Prediction Error}\label{subsec:exp_effect_pred}

In this experiment, we evaluate the effect prediction performance of the proposed transformer-based action encoder with a two-stage learning procedure, compared to alternative designs such as single-stage learning and a fully connected action encoder. All models are trained to predict an effect trajectory $\hat{e}_t$\, representing changes in object position, contact feedback, and force vectors at each timestep. We compute the mean absolute error between the predicted and ground truth positions of the object in effect trajectories on a test set of size $N = 1000$:

\begin{equation}
    Error = \frac{1}{NT} \sum_{i=1}^{N}  \sum_{t=1}^{T} ||\mathbf{\hat{e}}_t^{pos} -  \mathbf{e}_t^{pos}||_1
\end{equation}

The results are averaged over 5 runs with different random seeds. As shown in Table~\ref{tab:effect_error}, our proposed architecture, which combines two-stage learning with a transformer-based action encoder, achieves significantly lower prediction error than both the single-stage training variant (paired t-test, $p < 0.01$) and the fully-connected action encoder baseline (paired t-test, $p < 0.01$), validating our design choices.

\begin{table}[!t]
    \centering
    \caption{Mean Position Effect Prediction Errors averaged over 5 runs.}
    \label{tab:effect_error}
    \begin{tabular}{c c}
        \toprule
        Model & Error (cm)\\
        \midrule
        Effect Traj. Prediction Network (Ours) &  \textbf{1.70} $\pm$ \textbf{0.036} \\
        Effect Traj. Prediction Network (Single Stage)
        & 1.77 $\pm$ 0.041  \\ 
        Effect Traj. Prediction Network (Fully Connected) & 1.82 $\pm$ 0.058 \\
        \bottomrule
    \end{tabular}
    
\end{table}

\subsection{Planning Performance}\label{subsec:planning_performance}

\begin{figure}[!b]
    \centering
    \includegraphics[width=\linewidth]{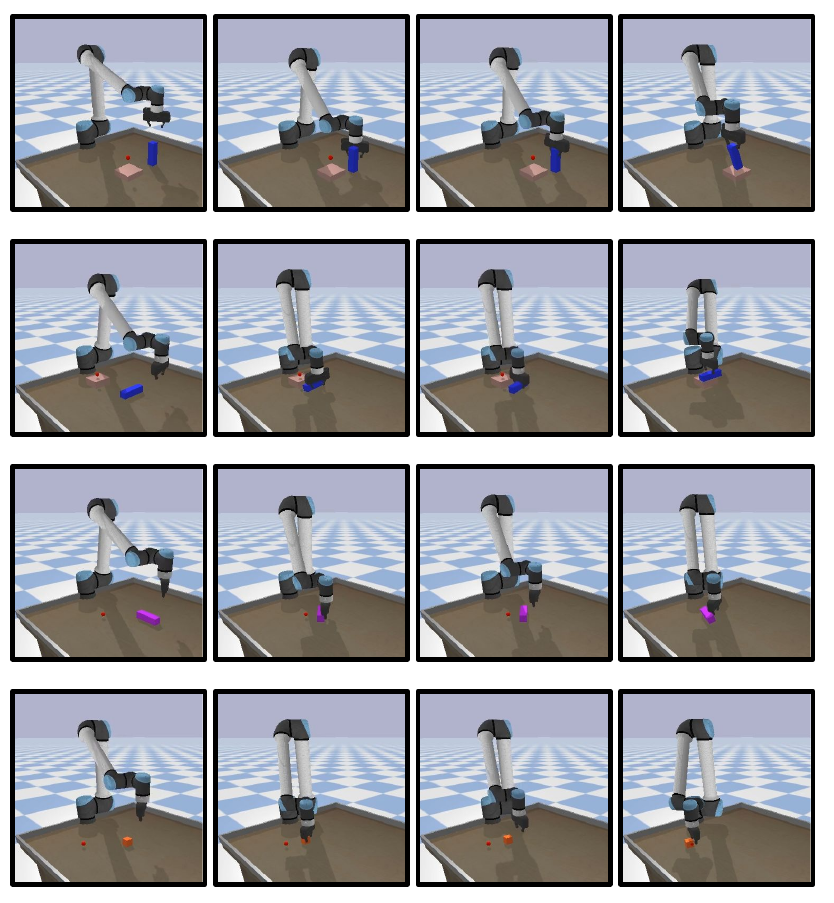}
    \caption{Planning examples of our approach for Block Z, Block Y, Block X, and Cube objects (from top to bottom). The red dots show the goal position in each case. The top two rows belong to Task~2, and the bottom two rows show Task~1.}
    \label{fig:planning_examples}
\end{figure}

In this experiment, we assess the planning performance of our method and the Diffusion Policy baseline on two tasks.
Task 1 is a tabletop task with no obstacles, in which the robot must move the object to a goal position within a larger area of the table. Task 2 is a stacking task where a height-varying block object is placed on the table, and the robot must stack the manipulated object on top of it. In general, the goal is not reachable with a single primitive execution, requiring multi-step planning with partial executions. Example planning executions of our approach can be seen in Fig.~\ref{fig:planning_examples}. To evaluate the performance in Task 1, we use the mean Euclidean distance error between the final object position and the goal with paired t-tests to assess statistical significance. For Task 2, we report the success rates and obtain significance values from McNemar's test~\cite{mcnemar1947note}.
We run each experiment 3 times and use 100 planning tests per task and object.

\subsubsection{Action Symbol Dimension Ablation}
\begin{figure}[!t]
    \centering
    \includegraphics[width=\linewidth]{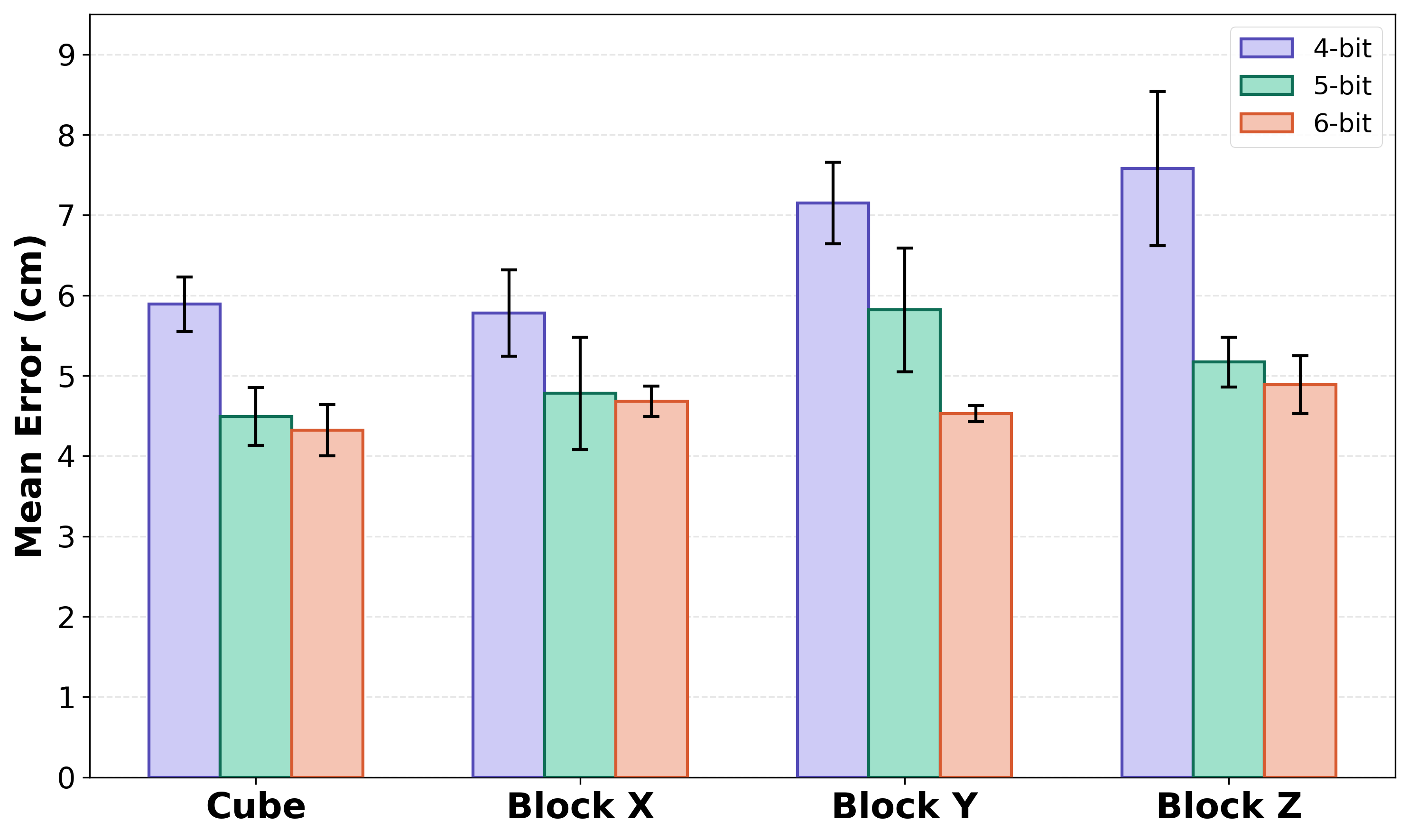}
    \caption{Mean planning errors (in cm) for Task 1 across different action symbol dimensions (4-bit, 5-bit, and 6-bit) for each object that is in the training set. The results are averaged over 3 runs, with 100 planning tests per object. Error bars refer to standard deviations across runs.}
    \label{fig:bit_ablations}
\end{figure}

We first performed an ablation study on the dimension of the action symbol to investigate its impact on planning performance. We train our entire pipeline with action symbol dimensions of 4, 5, and 6 bits and compute the mean planning errors for each configuration. As shown in Fig.~\ref{fig:bit_ablations},
the 4-bit configuration demonstrates poor performance compared to others due to limited planning resolution with only 16 distinct action possibilities. The 6-bit configuration achieves slightly lower planning errors than the 5-bit model. However, the planning time for the 6-bit configuration is higher since the branching factor in the search tree doubles with each additional bit. Therefore, we chose the 5-bit configuration for all subsequent experiments, as it provides a favorable trade-off between planning performance and computational cost.\\

\subsubsection{Planning Performance Comparison with Baseline}
\begin{table}[b]
    \centering
    \caption{Planning Results Across Seen Objects  for Both Tasks 1 \& 2}
    \label{tab:combined_plan_errors}
    
    \begin{tabular}{lcc}
        \toprule
        Object & Diffusion Policy & Ours \\
        \midrule
        \multicolumn{3}{c}{\textbf{Task 1 (Error in cm)}} \\
        \midrule
        Cube    & 5.09 $\pm$ 0.11 & \textbf{4.49 $\pm$ 0.36}  \\
        Block X & 6.50 $\pm$ 0.69 & \textbf{4.78 $\pm$ 0.70}  \\
        Block Y & 7.99 $\pm$ 0.45 & \textbf{5.82 $\pm$ 0.77}  \\
        Block Z & 7.05 $\pm$ 1.56 & \textbf{5.17 $\pm$ 0.31}  \\
        \midrule
        \multicolumn{3}{c}{\textbf{Task 2 (Success Rate $\%$)}} \\
        \midrule
        Cube    & 37.67 $\pm$ 3.3  & \textbf{55.33 $\pm$ 4.03} \\
        Block X & 17.33 $\pm$ 3.3  & \textbf{39.67 $\pm$ 5.44} \\
        Block Y & 19.67 $\pm$ 7.72 & \textbf{45.0 $\pm$ 2.16}  \\
        Block Z & 18.33 $\pm$ 2.87 & \textbf{69.67 $\pm$ 5.56} \\
        \bottomrule
    \end{tabular}
\end{table}

We evaluate both methods on cube, block X, block Y, and block Z objects for both Task 1 and Task 2. As reported in Table~\ref{tab:combined_plan_errors}, our proposed method achieves significantly lower planning errors across all objects in Task 1 (paired t-test, $p < 0.001$), and higher success rates in Task 2 (McNemar's test, $p < 0.001$). The performance of our method stems from the precision of search-based planning compared to goal-conditioned reactive policy execution. Furthermore, since our model learns the effects of actions at intermediate timesteps, it can generate new midway points that were not explicitly demonstrated in the training data. This capability provides superior in-distribution generalization compared to Diffusion Policy. The performance gap is specifically notable in Task 2 as stacking strictly requires
correct grasping of different objects, aligning with their grasp-affordances. For example, the direction of grasping differentiates Block X and Block Y as they require grasping from the narrower dimensions, and the grasping point differentiates Block Z, as it requires grasping from a higher point. The Diffusion Policy baseline failed to consistently match the grasp affordances of such objects, colliding with Block X and Block Y and overthrowing Block Z due to incorrect grasping. As a result, it performs poorly on this task, which requires precise execution of actions. On the other hand, our framework is preferable since it constructs object and action categories based on their effect trajectory, incentivizing representations that are influenced by both push and grasp actions in various directions.

\subsection{Few-Shot Object Generalization}\label{subsec:fewshot_exp}

\begin{figure}[!tb]
    \centering
    \includegraphics[width=\linewidth]{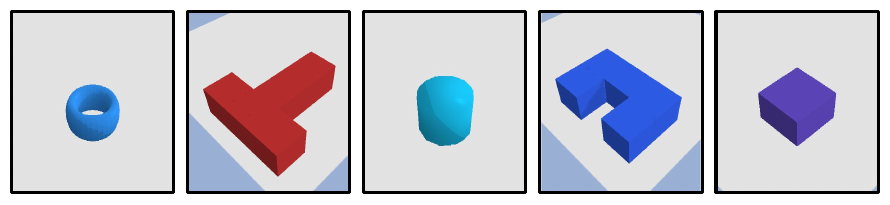}
    \caption{Novel objects for few-shot generalization. From left to right: Torus, T-Block, Cylinder, U-Block, and Wide Block.}
    \label{fig:unseen_objects}
\end{figure}

In our framework, we categorize objects based on similar effect-driven affordances rather than their visual appearances for planning. In this experiment, we introduce 5 new objects, including Torus, T-Block, Cylinder, U-Block, and Wide Block, to evaluate the impact of effect-driven categorization on generalization to novel objects. As shown in Fig.~\ref{fig:unseen_objects}, the novel objects have shapes or sizes that differ from those of the existing objects in the dataset. 

\begin{table}[!b]
    \centering
    \caption{Planning Results Across Unseen Objects for Both Tasks 1 \& 2}
    \label{tab:combined_generalization_errors}
    \begin{tabular}{lcc}
        \toprule
        Object & Diffusion Policy & Ours \\
        \midrule
        \multicolumn{3}{c}{\textbf{Task 1 (Error in cm)}} \\
        \midrule
        Wide Block & 5.28 $\pm$ 0.39 & \textbf{4.83 $\pm$ 0.11}\\
        Torus      & 5.57 $\pm$ 0.13 & \textbf{4.45 $\pm$ 0.26}\\
        U-Block    & 6.32 $\pm$ 0.84 & \textbf{4.76 $\pm$ 0.42}\\
        T-Block    & 6.84 $\pm$ 0.30 & \textbf{5.87 $\pm$ 0.33}\\
        Cylinder   & 6.96 $\pm$ 0.86 & \textbf{5.02 $\pm$ 0.42}\\
        \midrule
        \multicolumn{3}{c}{\textbf{Task 2 (Success Rate $\%$)}} \\
        \midrule
        Wide Block & 38.67 $\pm$ 3.77 & \textbf{44.33 $\pm$ 5.25} \\
        Torus      & 53.67 $\pm$ 3.30 & \textbf{56.33 $\pm$ 1.25} \\
        U-Block    & 27.0  $\pm$ 9.09 & \textbf{45.00 $\pm$ 4.24} \\
        T-Block    & 16.0  $\pm$ 0.82 & \textbf{31.33 $\pm$ 5.25} \\
        Cylinder   & 41.67 $\pm$ 6.13 & \textbf{43.00 $\pm$ 2.94}  \\
        \bottomrule
    \end{tabular}
\end{table}

\begin{figure}[tb]
    \centering
    \includegraphics[width=\linewidth]{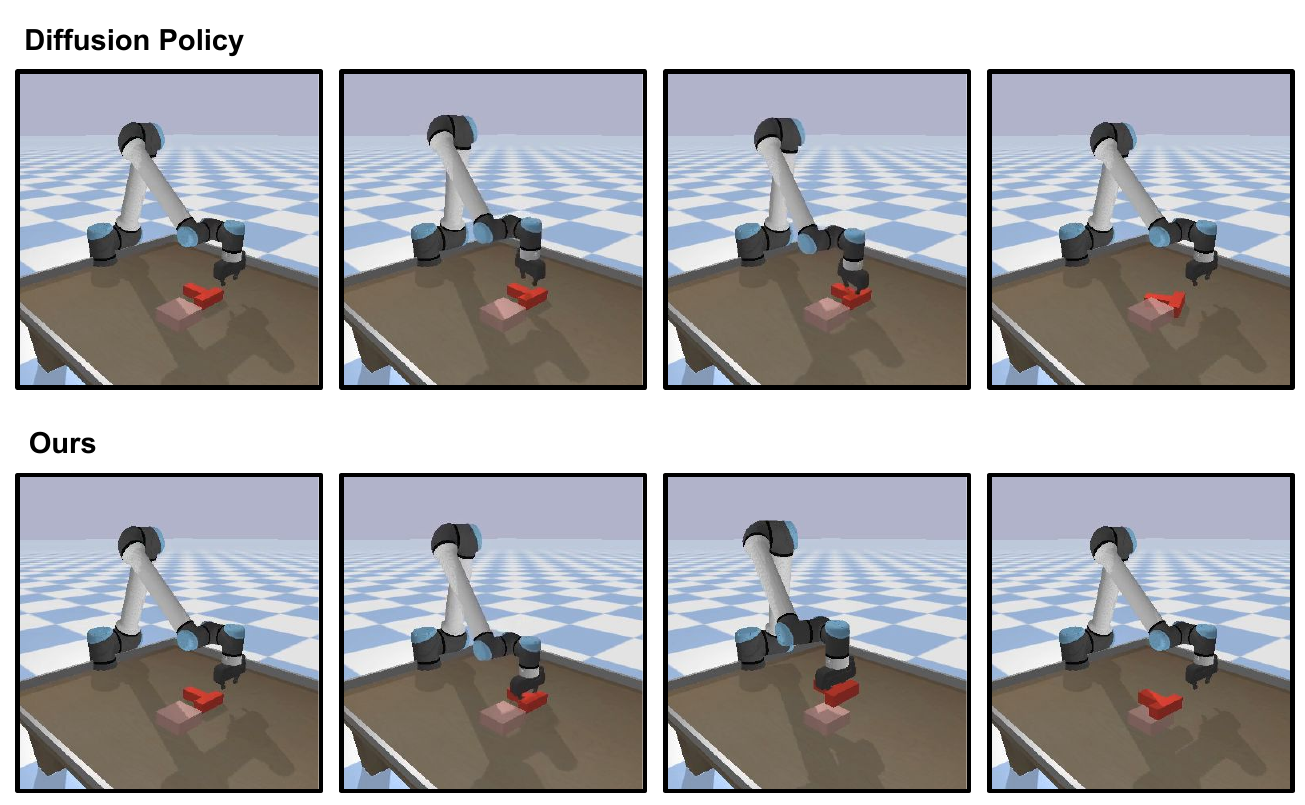}
    \caption{Example executions for stacking planning task including novel T-Block object, where Diffusion Policy (top) fails, while our proposed method (bottom) successfully performs the action.}
    \label{fig:fewshot_qualitative}
\end{figure}

Using our data collection procedure described in Section~\ref{subsec:data_collection}, we collect only three random demonstrations per novel object, and we then assign each an object symbol based on the lowest mean effect prediction error across the demonstrations, as explained in Section~\ref{sec:few-shot}.
We create 100 different planning tests for each novel object and task, and perform planning using the assigned object symbol in our proposed method. We compare our framework against the Diffusion Policy baseline, which is conditioned on the depth maps of the novel objects. As shown in Table \ref{tab:combined_generalization_errors}, our method significantly outperforms the Diffusion Policy baseline across all objects in both Task~1 (paired t-test, $p < 0.001$) and Task~2 (McNemar's test, $p < 0.001$). Figure~\ref{fig:fewshot_qualitative} illustrates a stacking action with the novel T-Block object, where the Diffusion Policy fails to grasp the object properly, while our method successfully picks and places it on top of the target.

\begin{figure}[!b]
    \centering
    \includegraphics[width=\linewidth]{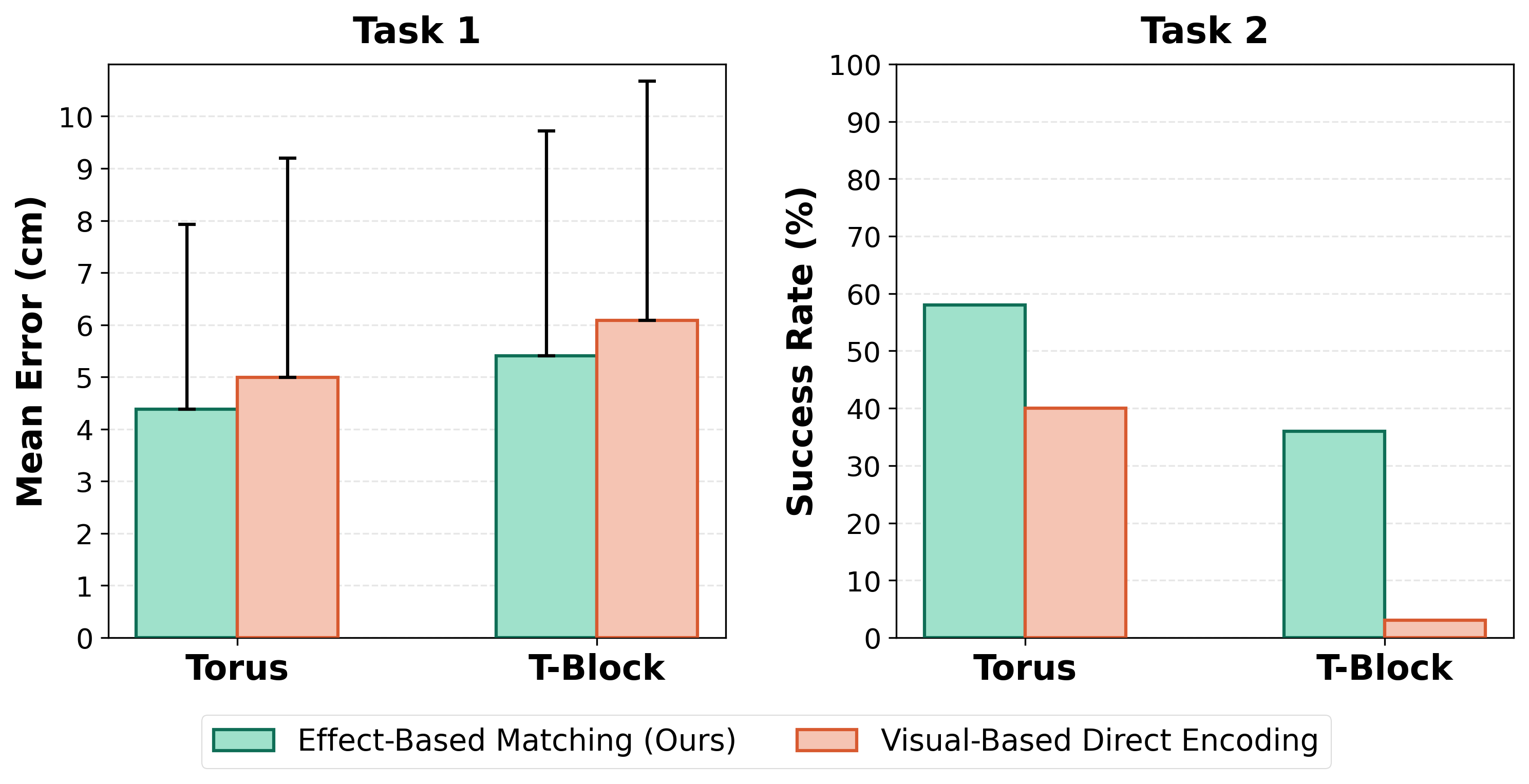}
    \caption{Comparison of effect-based matching (our proposed method) and visual-based encoding for novel object categorization on Torus and T-Block objects in both tasks, with 100 planning tests for each. For a visual-based encoding baseline, the symbol for the novel object is obtained directly from the object encoder without effect matching.}
    \label{fig:visual_comparison_plot}
\end{figure}

To further analyze the benefit of effect-based object categorization, we also modify our method such that the depth image of the novel object is directly passed through the trained object encoder, and the resulting symbol is used for planning without any effect-matching categorization. The encoder categorizes Torus as Ball, and T-Block as Block Y, while our effect-based categorization assigns Torus to Cube, and T-Block to Block X. These differences represent the distinctions in behavior of the objects that visual similarity fails to capture, such as Torus, which appears like a Ball but does not roll freely when push actions are applied, and T-Block is visually similar to both Block X and Block Y, but its grasp affordance aligns with Block X. As shown in Figure~\ref{fig:visual_comparison_plot}, our effect-based categorization yields lower planning errors for Task 1 (paired t-test, $p =0.05$), and significantly higher success rates for Task 2 (McNemar's test, $p < 0.001$) compared to visual-based alternative method. These results support the idea that object categorization based on behavioral effects, rather than image or geometric shapes, leads to more robust object generalization.

%% file: conclusion.tex
\section{Conclusion}

In this study, we propose a neuro-symbolic framework that discovers object and action symbols jointly from interaction data through multi-modal effect prediction and utilizes these learned symbols for multi-step planning on manipulation tasks. We utilize an encoder-decoder architecture with a binary bottleneck layer and a two-stage learning procedure, where the initial stage learns high-level action primitives from force and contact feedback, and the latter stage uses spatial effects to differentiate directional variations of primitives. In the experiments, we first validate our choices for the action encoder architecture and the two-stage learning approach. We then evaluate the planning performance of our method, which performs A* search over a discrete effect library with partial action executions and uses a symbol-conditioned low-level controller, outperforming the Diffusion Policy baseline significantly across all object types and manipulation tasks. Furthermore, we show that our effect-based object categorization enables robust object generalization with only a few demonstrations, outperforming a baseline that relies on visual appearance-based conditioning. Overall, we provide an approach to robotic manipulation planning that leverages symbolic abstractions derived from the multimodal effects of actions, enabling precise multi-step planning and efficient transfer for object generalization.

Our framework has several limitations that contain potential future research opportunities. Firstly, we focused on learning various unary single-object affordances and conducted multi-step planning for one object at a time. In the future, the discovery of inter-object relational symbols can be integrated into our system for both learning in complex environments and performing multi-object planning. In addition, the drift-based replanning mechanism we employ can trigger replanning only after an action is executed. This mechanism can be enhanced by incorporating a verification module that checks the feasibility of actions before execution. Lastly, we currently use the A* algorithm for planning, which lacks the speed and memory efficiency compared to the Planning Domain Definition Language (PDDL)~\cite{McDermott1998PDDL} solvers. In the future, our system can be developed to produce symbolic operators compatible with PDDL, enabling faster and more scalable symbolic planning.

%% file: ref.bib
@article{aktas2024multi,
  title={Multi-step planning with learned effects of partial action executions},
  author={Aktas, Hakan and Bozdogan, Utku and Ugur, Emre},
  journal={Advanced Robotics},
  volume={38},
  number={8},
  pages={562--576},
  year={2024},
  publisher={Taylor \& Francis}
}

@article{girgin2024multiobject,
  title={Multiobject Graph Affordance Network: Goal-Oriented Planning Through Learned Compound Object Affordances},
  author={Girgin, Tuba and U{\u{g}}ur, Emre},
  journal={IEEE Transactions on Cognitive and Developmental Systems},
  volume={17},
  number={4},
  pages={847--858},
  year={2024},
  publisher={IEEE}
}

@inproceedings{kilic2025predictability,
  author={Kilic, Burcu and Ahmetoglu, Alper and Ugur, Emre},
  booktitle={2025 IEEE International Conference on Development and Learning (ICDL)}, 
  title={Predictability-Based Curiosity-Guided Action Symbol Discovery}, 
  year={2025},
  volume={},
  number={},
  pages={1-6},
  doi={10.1109/ICDL63968.2025.11204386}}

@article{ahmetoglu2022deepsym,
  title={Deepsym: Deep symbol generation and rule learning for planning from unsupervised robot interaction},
  author={Ahmetoglu, Alper and Seker, M Yunus and Piater, Justus and Oztop, Erhan and Ugur, Emre},
  journal={Journal of Artificial Intelligence Research},
  volume={75},
  pages={709--745},
  year={2022}
}

@article{ahmetoglu2024discovering,
  title={Discovering predictive relational object symbols with symbolic attentive layers},
  author={Ahmetoglu, Alper and Celik, Batuhan and Oztop, Erhan and Ugur, Emre},
  journal={IEEE Robotics and Automation Letters},
  volume={9},
  number={2},
  pages={1977--1984},
  year={2024},
  publisher={IEEE}
}

@article{ahmetoglu2025symbolic,
  title={Symbolic manipulation planning with discovered object and relational predicates},
  author={Ahmetoglu, Alper and Oztop, Erhan and Ugur, Emre},
  journal={IEEE Robotics and Automation Letters},
  year={2025},
  publisher={IEEE}
}

@article{maddison2016concrete,
  title={The concrete distribution: A continuous relaxation of discrete random variables},
  author={Maddison, Chris J and Mnih, Andriy and Teh, Yee Whye},
  journal={arXiv preprint arXiv:1611.00712},
  year={2016}
}

@article{jang2016categorical,
  title={Categorical reparameterization with gumbel-softmax},
  author={Jang, Eric and Gu, Shixiang and Poole, Ben},
  journal={arXiv preprint arXiv:1611.01144},
  year={2016}
}

@misc{kingma2013auto,
  title={Auto-encoding variational bayes},
  author={Kingma, Diederik P and Welling, Max and others},
  year={2013},
  publisher={Banff, Canada}
}

@INPROCEEDINGS{Ugur-RSS-19,
    AUTHOR    = {Muhammet Yunus Seker AND Mert Imre AND Justus Piater AND Emre Ugur}, 
    TITLE     = {Conditional Neural Movement Primitives}, 
    BOOKTITLE = {Proceedings of Robotics: Science and Systems}, 
    YEAR      = {2019}, 
    ADDRESS   = {FreiburgimBreisgau, Germany}, 
    MONTH     = {June}, 
    DOI       = {10.15607/RSS.2019.XV.071}}

@article{vaswani2017attention,
  title={Attention is all you need},
  author={Vaswani, Ashish and Shazeer, Noam and Parmar, Niki and Uszkoreit, Jakob and Jones, Llion and Gomez, Aidan N and Kaiser, {\L}ukasz and Polosukhin, Illia},
  journal={Advances in neural information processing systems},
  volume={30},
  year={2017}
}

@article{kingma2014adam,
  title={Adam: A method for stochastic optimization},
  author={Kingma, Diederik P and Ba, Jimmy},
  journal={arXiv preprint arXiv:1412.6980},
  year={2014}
}

@article{ugur2025,
  title={Neuro-Symbolic Robotics},
  author={Emre Ugur and Alper Ahmetoglu and Yukie Nagai and Tadahiro Taniguchi and Matteo Saveriano and Erhan Oztop},
  note  = {\url{http://dx.doi.org/10.13140/RG.2.2.25854.09283}},
  year={2025}
}

@inproceedings{ugur2015bottom,
  title={Bottom-up learning of object categories, action effects and logical rules: From continuous manipulative exploration to symbolic planning},
  author={Ugur, Emre and Piater, Justus},
  booktitle={2015 IEEE International Conference on Robotics and Automation (ICRA)},
  pages={2627--2633},
  year={2015},
  organization={IEEE}
}

@article{tekden2024object,
  title={Object and relation centric representations for push effect prediction},
  author={Tekden, Ahmet E and Erdem, Aykut and Erdem, Erkut and Asfour, Tamim and Ugur, Emre},
  journal={Robotics and Autonomous Systems},
  volume={174},
  pages={104632},
  year={2024},
  publisher={Elsevier}
}

@article{konidaris2018skills,
  title={From skills to symbols: Learning symbolic representations for abstract high-level planning},
  author={Konidaris, George and Kaelbling, Leslie Pack and Lozano-Perez, Tomas},
  journal={Journal of Artificial Intelligence Research},
  volume={61},
  pages={215--289},
  year={2018}
}

@article{konidaris2019necessity,
  title={On the necessity of abstraction},
  author={Konidaris, George},
  journal={Current opinion in behavioral sciences},
  volume={29},
  pages={1--7},
  year={2019},
  publisher={Elsevier}
}

@inproceedings{asai2018classical,
  title={Classical planning in deep latent space: Bridging the subsymbolic-symbolic boundary},
  author={Asai, Masataro and Fukunaga, Alex},
  booktitle={AAAI},
  volume={32},
  number={1},
  year={2018}
}

@article{asai2022classical,
  title={Classical planning in deep latent space},
  author={Asai, Masataro and Kajino, Hiroshi and Fukunaga, Alex and Muise, Christian},
  journal={Journal of Artificial Intelligence Research},
  volume={74},
  pages={1599--1686},
  year={2022}
}

@inproceedings{asai2021learning,
author = {Asai, Masataro and Muise, Christian},
title = {Learning neural-symbolic descriptive planning models via cube-space priors: the voyage home (to STRIPS)},
year = {2021},
isbn = {9780999241165},
booktitle = {Proceedings of the Twenty-Ninth International Joint Conference on Artificial Intelligence},
articleno = {371},
numpages = {7},
location = {Yokohama, Yokohama, Japan},
series = {IJCAI'20}
}

@incollection{gibson1977theory,
  title={The theory of affordances},
  author={Gibson, James J},
  booktitle={Perceiving, Acting, and Knowing: Toward an Ecological Psychology},
  editor={Shaw, Robert E. and Bransford, John},
  pages={67--82},
  year={1977},
  publisher={Lawrence Erlbaum Associates},
  address={Hillsdale, NJ}
}

@article{jamone2018affordances,
  title={Affordances in psychology, neuroscience, and robotics: A survey},
  author={Jamone, Lorenzo and Ugur, Emre and Cangelosi, Angelo and Fadiga, Luciano and Bernardino, Alexandre and Piater, Justus and Santos-Victor, Jos{\'e}},
  journal={IEEE Transactions on Cognitive and Developmental Systems},
  volume={10},
  number={1},
  pages={4--25},
  year={2018},
  publisher={IEEE}
}

@article{sun2000symbol,
  title={Symbol grounding: a new look at an old idea},
  author={Sun, Ron},
  journal={Philosophical Psychology},
  volume={13},
  number={2},
  pages={149--172},
  year={2000},
  publisher={Taylor \& Francis}
}

@article{gibson1984development,
 ISSN = {00093920, 14678624},
 author = {Eleanor J. Gibson and Arlene S. Walker},
 journal = {Child Development},
 number = {2},
 pages = {453--460},
 publisher = {[Wiley, Society for Research in Child Development]},
 title = {Development of Knowledge of Visual-Tactual Affordances of Substance},
 urldate = {2026-02-16},
 volume = {55},
 year = {1984}
}

@article{smith2005action,
  title={Action alters shape categories},
  author={Smith, Linda B},
  journal={Cognitive Science},
  volume={29},
  number={4},
  pages={665--679},
  year={2005},
  publisher={Wiley Online Library}
}

@article{asada2009cognitive,
  title={Cognitive developmental robotics: A survey},
  author={Asada, Minoru and Hosoda, Koh and Kuniyoshi, Yasuo and Ishiguro, Hiroshi and Inui, Toshio and Yoshikawa, Yuichiro and Ogino, Masaki and Yoshida, Chisato},
  journal={IEEE transactions on autonomous mental development},
  volume={1},
  number={1},
  pages={12--34},
  year={2009},
  publisher={IEEE}
}

@article{oztop2004infant,
  title={Infant grasp learning: a computational model},
  author={Oztop, Erhan and Bradley, Nina S. and Arbib, Michael A.},
  journal={Experimental Brain Research},
  volume={158},
  number={4},
  pages={480--503},
  year={2004},
  publisher={Springer},
  doi={10.1007/s00221-004-1914-1}
}

@inproceedings{kroemer2022search,
  title={Search-based task planning with learned skill effect models for lifelong robotic manipulation},
  author={Liang, Jacky and Sharma, Mohit and LaGrassa, Alex and Vats, Shivam and Saxena, Saumya and Kroemer, Oliver},
  booktitle={2022 International Conference on Robotics and Automation (ICRA)},
  pages={6351--6357},
  year={2022},
  organization={IEEE}
}

@article{li2021planning,
  title={Planning in learned latent action spaces for generalizable legged locomotion},
  author={Li, Tianyu and Calandra, Roberto and Pathak, Deepak and Tian, Yuandong and Meier, Franziska and Rai, Akshara},
  journal={IEEE Robotics and Automation Letters},
  volume={6},
  number={2},
  pages={2682--2689},
  year={2021},
  publisher={IEEE}
}

@article{ning2023where2explore,
  title={Where2explore: Few-shot affordance learning for unseen novel categories of articulated objects},
  author={Ning, Chuanruo and Wu, Ruihai and Lu, Haoran and Mo, Kaichun and Dong, Hao},
  journal={Advances in Neural Information Processing Systems},
  volume={36},
  pages={4585--4596},
  year={2023}
}

@inproceedings{
lorang2025fewshot,
title={Few-Shot Neuro-Symbolic Imitation Learning for Long-Horizon Planning and Acting},
author={Pierrick Lorang and Hong Lu and Johannes Huemer and Patrik Zips and Matthias Scheutz},
booktitle={9th Annual Conference on Robot Learning},
year={2025},
}

@inproceedings{wang2022adaafford,
  title={Adaafford: Learning to adapt manipulation affordance for 3d articulated objects via few-shot interactions},
  author={Wang, Yian and Wu, Ruihai and Mo, Kaichun and Ke, Jiaqi and Fan, Qingnan and Guibas, Leonidas J and Dong, Hao},
  booktitle={European conference on computer vision},
  pages={90--107},
  year={2022},
  organization={Springer}
}

@inproceedings{
tian2025oafford,
title={{O$^3$Afford}: One-Shot 3D Object-to-Object Affordance Grounding for Generalizable Robotic Manipulation},
author={Tongxuan Tian and Xuhui Kang and Yen-Ling Kuo},
booktitle={9th Annual Conference on Robot Learning},
year={2025},
}

@article{chi2024diffusionpolicy,
	author = {Cheng Chi and Zhenjia Xu and Siyuan Feng and Eric Cousineau and Yilun Du and Benjamin Burchfiel and Russ Tedrake and Shuran Song},
	title ={Diffusion Policy: Visuomotor Policy Learning via Action Diffusion},
	journal = {The International Journal of Robotics Research},
	year = {2024},
}

@MISC{coumans2021,
author =   {Erwin Coumans and Yunfei Bai},
title =    {PyBullet, a Python module for physics simulation for games, robotics and machine learning},
howpublished = {\url{http://pybullet.org}},
year = {2016--2021}
}

@article{mcnemar1947note,
  title={Note on the sampling error of the difference between correlated proportions or percentages},
  author={McNemar, Quinn},
  journal={Psychometrika},
  volume={12},
  number={2},
  pages={153--157},
  year={1947},
  publisher={Springer-Verlag}
}

@techreport{McDermott1998PDDL,
  author      = {McDermott, Drew and Ghallab, Malik and Howe, Adele and Knoblock, Craig and Ram, Ashwin and Veloso, Manuela and Weld, Daniel and Wilkins, David},
  title       = {{PDDL}---The Planning Domain Definition Language},
  institution = {Yale Center for Computational Vision and Control},
  number      = {TR-98-003},
  year        = {1998},
  type        = {Technical Report}
}
